\documentclass[final]{cvpr}

\usepackage[utf8]{inputenc} 
\usepackage[T1]{fontenc}    
\usepackage{times}
\usepackage{epsfig}
\usepackage{graphicx}
\usepackage{amsmath}
\usepackage{amssymb}
\usepackage{setspace}
\usepackage[T1]{fontenc}    
\usepackage{url}            
\usepackage{booktabs}       
\usepackage{amsfonts}       
\usepackage{nicefrac}       
\usepackage{microtype}      

\usepackage{amsmath}
\usepackage{csquotes}

\usepackage{amsthm}
\usepackage{float}

\usepackage{subfigure}
\usepackage{epsfig}
\usepackage{epstopdf}
\usepackage{color,xcolor}
\usepackage[ruled,linesnumbered]{algorithm2e}
\usepackage{cite}
\usepackage{bm}
\usepackage{multirow}
\usepackage{verbatim}
\usepackage{flushend}

\newcommand{\diag}{\mathrm{diag}}

\usepackage[pagebackref=true,breaklinks=true,colorlinks,bookmarks=false]{hyperref}

\def\cvprPaperID{3689} 
\def\confYear{CVPR 2021}

\begin{document}

\title{Image Denoising by Gaussian Patch Mixture Model and Low Rank Patches}

\author{Jing Guo\textsuperscript{\rm 1},~  
        Shuping Wang\textsuperscript{\rm 1},~ 
        Chen Luo\textsuperscript{\rm 1},~ 
        Qiyu~Jin\textsuperscript{\rm 1},~
        Michael Kwok-Po Ng\textsuperscript{\rm 2},~ \\
        \textsuperscript{\rm 1} School of Mathematical Science, Inner Mongolia University, Hohhot, China\\
        \textsuperscript{\rm 2} Department of Mathematics, University of Hong Kong, Pokfulam, Hong Kong , China\\
{\tt\small guojing19@aliyun.com,  wangsp1993@163.com, luochen\_2019@163.com, }\\
{\tt\small qyjin2015@aliyun.com,  mng@maths.hku.hk}
}

\maketitle

\begin{abstract}
Non-local self-similarity based low rank algorithms are the state-of-the-art methods for image denoising. In this paper, a new method is proposed by solving two issues: how to improve similar patches matching accuracy and build an appropriate low rank matrix approximation model for Gaussian noise. For the first issue, similar patches can be found locally or globally. Local patch matching is to find similar patches in a large neighborhood which can alleviate noise effect, but the number of patches may be insufficient. Global patch matching is to determine enough similar patches but the error rate of patch matching may be higher. Based on this, we first use local patch matching method to reduce noise and then use Gaussian patch mixture model to achieve global patch matching. The second issue is that there is no low rank matrix approximation model to adapt to Gaussian noise. We build a new model according to the characteristics of Gaussian noise, then prove that there is a globally optimal solution of the model. By solving the two issues, experimental results are reported to show that the proposed approach outperforms the state-of-the-art denoising methods includes several deep learning ones in both PSNR / SSIM values and visual quality.
\end{abstract}

\section{Introduction}

As a classical and fundamental problem in low level vision, image denoising has been extensively explored and it is still under investigation as well in methodological aspects. In general, image denoising aims to recover the latent noise-free image from its noise-corrupted version.
During the past decades, image denoising has got great development in a variety of models including spatial filtering ~\cite{Tomasi1998Bilateral},  total variation~\cite{Rudin1992Nonlinear}, wavelet/curvelet transform~\cite{donoho1995noising}, noise detector~\cite{cai2007minimization,chan2010efficient,dong2007detection}, sparse representation~\cite{cai2014data,dong2012nonlocal}, non-local means and its varieties ~\cite{buades2005non,chen2015external,dabov2007image,gu2014weighted,zhang2018kernel}, deep learning\cite{Chen2018Image,He2016Deep,simonyan2014very,zhang2017beyond,Zhang2018FFDNet}, etc.

The landmark work of image denoising is non-local means (NLM)~\cite{buades2005non} which brings a new era by finding non-local similar patches within the image.
Quite a few patch-based denoising methods~\cite{chen2015external,dabov2007image,dong2012nonlocal,gu2014weighted,wang2019seismic,Zhang2018FFDNet} achieve great success by exploiting the internal self-similarity prior.
A quintessential example should be cited that BM3D~\cite{dabov2007image} uses 3D filtering in transformed domain by patch matching,
which is quite effective and has been a benchmark in image denoising.
Here, patch matching is performed within a relatively large local window instead
of globally, and consequently the number of patches may not be sufficient that leads to ringing artifacts.
In order to improve the accuracy of patch matching, external natural patch priors are used to regularize the denoising process.
Method~\cite{chatterjee2010learning} learns the covariance matrix and the measure of redundancy for each patch based on clustered natural noise-free patch database.
Zoran and Weiss~\cite{Zoran2011From} gave high likelihood values for patches sampled from clean images using Gaussian mixture models (GMM), and reconstructed the latent image by maximizing the expected patch log likelihood.
Though external patch priors based approaches have achieved remarkable success in image restoration, they are still local and do not make full use of image internal self-similarity.
Some methods ~\cite{dong2012nonlocal,gu2014weighted} globally search for the most similar patches across the whole image and have demonstrated competitive results.
Though global patch matching is to determine enough similar patches, the error rate of patch matching may be higher, which results in destroying image details unsatisfactorily.
Methods PCLR~\cite{chen2015external} and PGPD~\cite{Xu2016Patch} integrate external patch priors and internal self-similarity into one framework.
They learn GMM prior information from noise-free image patches to guide the noisy patches classification and the subsequent low rank subspace learning, which makes the latent patch reconstruction more robust to noise.
However, there is no obvious improvement in denoising performance,  compared with local patch matching methods.
We argue that such utilization of GMM prior information is not effective enough because it neglects the noise of degraded image, which interferes with the selection of similar patches.
How to precisely select similar patches is still an open problem.

Besides,
low rank matrix approximation (LRMA), which aims to
recover the underlying low rank matrix from its degraded
observation, 
appears in a wide range of applications in  various fields including computer vision,
machine learning, signal processing and bioinformatics.
For instance, use of low rank approximation can be found in applications such as 
face recognition~\cite{gao2017learning,dong2019low}, 
background modeling and foreground object detection~\cite{gu2014weighted,javed2016spatiotemporal,erichson2019compressed}, 
image alignment~\cite{peng2012rasl,zhang2019robust},
video denoising~\cite{ji2010robust,ren2020lr3m} image restoration~\cite{chen2015external,wang2004image,dong2012nonlocal}, among others. 
Since direct rank minimization is difficult to solve, the problem
is generally relaxed by substitutively minimizing the nuclear norm of the estimated matrix~\cite{fazel2003matrix}. 
The nuclear norm of a matrix $\mathbf{X}\in \mathbb{R}^{m\times n}$, denoted by $\| \mathbf{X} \|_{*}$, is defined as the sum of its singular values, 
i.e. $\|\mathbf{X}\|_{*} = \sum_{i}^{n}\lambda_{\mathbf{X},i} $, 
where $\lambda_{\mathbf{X},i}$ denotes the $i$-th singular value of $\mathbf{X}$.
The nuclear norm minimization  model has been attracting significant attention due to its rapid development in both theory and implementation.
 Cai et al.~\cite{cai2010singular} proved that the  model can be easily solved in a closed form by imposing a soft thresholding operation on the singular values of the observation matrix.
 Candès et al.~\cite{candes2011robust} proved that from the noisy input, its intrinsic low-rank reconstruction can be exactly achieved with a high probability by the model.
 Many state-of-the-art denoising methods
 ~\cite{ge2016structure,wu2017structure,lin2011linearized,dong2012nonlocal}  have been proposed in
recent years by utilizing nuclear norm proximal as the key proximal technique.
Albeit its success as mentioned before, nuclear norm minimization  model still has certain limitations. 
Gu et al.~\cite{gu2014weighted} proposed weighted nuclear norm minimization model
trying to achieve a better estimation of the latent data from the
corrupted input.
The weighted nuclear norm of a matrix  $\mathbf{X}\in \mathbb{R}^{m\times n}$  is given by $\| \mathbf{X} \|_{\mathbf{w},*}$,  
where $\mathbf{w} = (w_1,w_{2},\cdots,w_n)^T$ and
$w_i\geq 0$ is a non-negative weight assigned to $\lambda_{\mathbf{X},i}$.
The nuclear norm minimization model provides superior performance and efficiency for matrix completion or impulse noise. However, they ignore the difference between Gaussian noise and impulse noise,
then building a low rank model for Gaussian noise  is still a worthwhile work.

This article will address two issues: (1) how to improve similar patches matching accuracy; (2) build an appropriate low rank matrix approximation model for Gaussian noise.

Firstly, unlike PCLR~\cite{chen2015external} and PGPD~\cite{Xu2016Patch} clustering the noisy patches directly, we first carry out a preliminary denoising of the image by a local patch matching based denoising algorithm and get a preprocessed image, then cluster all preprocessed patches into $K$ classes. If a class contains too many similar patches, we use k-means to divide the big class into some smaller ones. We next record all patches location information. Consequently,
in order to classify noisy patches, we  replace all preprocessed patches by the corresponding noisy ones in each class according to the location information.
Furthermore, the preprocessed image is updated in each iteration in order to improve patch matching accuracy.
This idea significantly increases the accuracy of similar patch clustering. Here, we use a local patch matching based denoising method because it matches similar patches with higher precision. Since BM3D~\cite{dabov2007image} is the state-of-the-art local patch matching based method and a real-time computing algorithm, we use it as a preprocessing algorithm in this paper.
Though some global patch matching based methods outperform BM3D, they introduce some errors for patch matching and the errors can't be corrected in the future denoising steps.
In addition, the methods PCLR~\cite{chen2015external} and PGPD~\cite{Xu2016Patch} are iterative algorithms.
This means that the methods PCLR~\cite{chen2015external} and PGPD~\cite{Xu2016Patch} have already used a global patch matching based algorithm (the $(t-1)$-th step denoising) as a preprocessing algorithm for the $t$-th step denoising.

Secondly, according to the characteristics of Gaussian noise, the minimization of $\| \mathbf{X}\|^{2}_{F}$ i.e. the square of the Frobenius norm of $\mathbf{X}$ adapts Gaussian noise model. Inspired by~\cite{zhang2018kernel} and the equation $\|\mathbf{X}\|_{F}^{2} = \sum_{i} \lambda_{\mathbf{X},i}^{2} = \|\mathbf{X}\mathbf{X}^{T}\|_{*}$,  similar to nuclear norm minimization model~\cite{cai2010singular}, we will establish Gaussian nuclear norm minimization model by minimizing $\|\mathbf{X}\mathbf{X}^{T}\|_{*}$  and then efficiently solve the problem.

As noted above, the proposed method uses GMM to achieve both local and global patch matching, and the mixture of patches can be further regularized via k-means clustering procedure and low rank minimization. This process may seem complicated, but it is very effective. In the low rank denoising part, we use the soft threshold given by~\cite{gu2014weighted} which is good to smooth the image. The proposed model is solved by an alternating minimization procedure and it converges rapidly.

The rest of the paper is organized as follows. In Section \ref{sec 2}, we propose our algorithm by using GMM so that both local and global patch matching can be achieved. The complexity and convergence of the algorithm are also been discussed. Section \ref{sec 3} shows experiment results and discussion. Conclusions and future works are given in Section \ref{sec 4}.

\section{The Proposed Algorithm}
\label{sec 2}

Throughout this paper, we denote scalars, vectors and matrices by non boldfaced, boldfaced lower-case, and boldfaced upper-case letters, respectively.

\subsection{The low rank regularization model based on GMM prior}
The noise-corrupted images  are often
represented by the following simplified model:
\begin{eqnarray}
  \mathbf{Y}= \mathbf{X} + \bm{\varepsilon },
  \label{Model AWG}
\end{eqnarray}
where $\mathbf{Y} \in \mathbb{R}^{m\times n}$ is noisy image, $\mathbf{X}\in \mathbb{R}^{m\times n}$ is latent original image and $\bm{\varepsilon }\in \mathbb{R}^{m\times n}$ is additive white Gaussian (AWG) noise  matrix of mean $0$ and standard deviation $\sigma$.
Let $\mathbf{y}_{i}$ and $\mathbf{x}_{i}$ be the vectorized $s \times s$ image patches   which are extracted from the noisy image $\mathbf{Y}$ and latent original image $\mathbf{X}$ at the $i$-th
pixel, respectively.
All noisy patches are  divided into $K$ similar patch groups.
Consequently, these similar patches in $k$-th group are collected to form a reshaped matrix $\mathbf{Y}_{k}=[\mathbf{y}_{k,1},\mathbf{y}_{k,2},\cdots,\mathbf{y}_{k,q_{k}}]^{T} \in \mathbb{R} ^{q_{k} \times s^2}$, where $q_{k}$ may be different for each $k$.
The noise model \eqref{Model AWG} can be rewritten in the patch-based representation as follow:
\begin{eqnarray}
    \mathbf{Y}_{k}= \mathbf{X}_{k} + \bm{\varepsilon }_{k},\quad k=1,2,\cdots,K,
    \label{Model AWGPatch}
\end{eqnarray}
where $\mathbf{Y}_{k},\mathbf{X}_{k},\bm{\varepsilon }_{k}\in \mathbb{R} ^{q_{k} \times s^2}$ denote the patch reshaped  matrices from the noisy image $\mathbf{Y} $, original image $\mathbf{X}$ and noise matrix $\bm{\varepsilon }$,  respectively.

In this paper,  we will introduce preprocessed image based GMM prior low rank regularization model (PG-LR):

\begin{eqnarray} \label{Model ours}
  &&(  \widetilde {\mathbf{X}}, \widetilde {\mathbf{c}}, \{ \widetilde {\mathbf{X}}_{k}\} ) =  \quad \mathop {\arg \min }\limits_{ \mathbf{X},\mathbf{c}, \{ {\mathbf{X}_k}\} }  
  \left\| {\mathbf{Y} - \mathbf{X}} \right\|_F    ^2   
  \nonumber\\
    &&\quad
    + 
    \sum\limits_{k = 1}^K \left( \left\| \mathbf{Y}_{k}\mathbf{Y}_{k} ^{T}- \mathbf{X}_{k}\mathbf{X}_{k}^{T} \right\|_F^2 + q_{k}\sigma^{2}\left\| \mathbf{X}_{k}\mathbf{X}_{k}^{T} \right\|_ * \right)\nonumber\\
    &&\quad - \sum\limits_{{ {i}} = 1}^{ {m\times n}} {\log p({\mathbf{x}^{pr}_{i}}, {{{c}}_{{i}}}} \left| \bm{\Theta}  \right.),
\end{eqnarray}
 where$\|\cdot\|_{F}$ is the Frobenius norm, 
 $\|\cdot\|_{*}$ denotes nuclear norm.  $\sum\limits_{{{i}} = 1}^{ {m\times n}} {\log p({\mathbf{x}^{pr}_{i}}, {{{c}}_{{i}}}} \left| \Theta  \right.)$ is  preprocessed  image based  patch cluster log likelihood term, $\mathbf{x}^{pr}_{i}$ is the $i$-th patch extracted from preprocessed image $\mathbf{X}^{pr}$, $\mathbf{c} = (c_1, c_2, \cdots, c_{m\times n})$ and $c_i\in \{1,2,\cdots,K\}$  denotes which class $\mathbf{x}^{pr}_{i}$ is from. 
 Our model effectively incorporates image internal self-similarity and external patch priors into the low rank denoising model. There are two advantages of such clustering based low rank regularization.
 
 First,  the Gaussian patch mixture model based patch clustering can take the advantages of local and global patch matching. Local patch matching is to find similar patches in a large neighborhood which can alleviate noise effect, while the number of patches may not be sufficient. Global patch matching aims to determine enough similar patches but the error rate of patch matching may be higher. The model first use local patch matching method to reduce noise and then use Gaussian patch mixture model to achieve global patch matching. The detail will be discussed in Subsection \ref{Sec: GPMM}.

Second, the nuclear norm minimization model 
\cite{cai2010singular} is an outstanding model for removing impulse noise and inpainting, but not adapt to Gaussian noise. Here according to the statistic properties of Gaussian noise and the nuclear norm minimization model \cite{cai2010singular}, we build a new model for Gaussian noise: $\min_{\mathbf{X}_k}\left\| \mathbf{Y}_{k}\mathbf{Y}_{k} ^{T}- \mathbf{X}_{k}\mathbf{X}_{k}^{T} \right\|_F^2 + q_{k}\sigma^{2}\left\| \mathbf{X}_{k}\mathbf{X}_{k}^{T} \right\|_ *$, where $ q_{k}$ is the number of rows in the matrix $\mathbf{X}_{k}$. We prove that there is a global optimal solution of the model and the simulation experiment shows that our model is much better than the nuclear norm minimization model \cite{cai2010singular} for Gaussian noise (see the supplementary material). Subsection \ref{Sec: gnnm} will give details.




\subsection{Improved Gaussian Patch Mixture Model}
\label{Sec: GPMM}
In this subsection, we first recall GMM model~\cite{Zoran2011From}. All noise-free image patches $\mathbf{x}_i$,  ($i=1,2,\cdots,m\times n$) are divided into
$K$ groups with a parameter $\bm{\theta}_{k}$, $k=1,2,\cdots,K$, which describes a Gaussian
density function parameterized by the mean $\bm{\mu}_{k}$ and covariance matrix $\bm{\Sigma}_{k}$.
Thus, the probability of each patch $\mathbf{x}_i$ could be defined as a weighted sum of $K$-Gaussians:
\begin{eqnarray}
    p(\mathbf{x}_i|\bm{\Theta}) = \sum_{k=1}^{K} \omega_{k}p_{k}(\mathbf{x}_i|\bm{\theta}_{k}),
    \label{Model GMM}
\end{eqnarray}
where
\begin{eqnarray}
    p_{k}(\mathbf{x}_i|\bm{\theta}_{k}) \propto \exp\left\{-\frac{1}{2}(\mathbf{x}_i -\bm{\mu}_{k} )^{T}\bm{\Sigma}_{k}^{-1} (\mathbf{x}_i -\bm{\mu}_{k} )\right\},
    \label{Gaussian function}
\end{eqnarray}
and
$(\mathbf{x}_i -\bm{\mu}_{k} )^{T}$
is the transpose of
$(\mathbf{x}_i -\bm{\mu}_{k} )$,
and $\bm{\Theta} =(\omega_{1},\omega_{2},\cdots,\omega_{K},\bm{\theta}_{1},\bm{\theta}_{2},\cdots,\bm{\theta}_{K})$ is the set of parameters with $\sum_{k=1}^{K} \omega_{k} =1$ and $\omega_{k}\geq 0$, $k=1,2,\cdots,K$.

We investigate the model (\ref{Model GMM}) and discover that $\mathbf{x}_i$ is unknown latent image patch and consequently it's impossible to calculate $p(\mathbf{x}_i|\bm{\Theta})$. In the model (\ref{Model GMM}), it normally takes noisy patch $\mathbf{y}_i$ instead of noise-free patch $\mathbf{x}_i$. As a result, the error rate of patch clustering increases significantly due to noise. In this paper, we will solve the problem.

The latent image patch $\mathbf{x}_i$ is unknown and the noisy image patch $\mathbf{y}_i$ is unfavorable, then we should find another solution.
There are quite a few excellent denoising algorithms~\cite{buades2005non,chen2015external,dabov2007image,gu2014weighted}, and a distinguished  one is BM3D~\cite{dabov2007image}. BM3D is a real-time computing algorithm and gets favorable restoration results. Furthermore, it matches similar patches in a large neighborhood, which alleviates noise effect for selection of similar patches. We take BM3D as a image preprocessing algorithm and get a preprocessed image $\mathbf{X}^{pr}$ that is a rosy estimate of the noise-free one.
For this reason, the preprocessed image based patch cluster log likelihood (P-CLL) term is given as follow:
\begin{eqnarray}
   \log \prod _{i=1}^{m\times n}p(\mathbf{x}_{i}^{pr},c_{i}|\bm{\Theta}) =  \sum\limits_{{ {i}} = 1}^{ {m\times n}} {\log p({\mathbf{x}^{pr}_{i}}, {{{c}}_{{i}}}} \left| \bm{\Theta}  \right.).
\end{eqnarray}
Obviously, our model is feasible and efficient  as we know $\mathbf{x}^{pr}_{i}$  is the pleasurable estimate of $\mathbf{x}_{i}$.
Consequently, we cluster all preprocessed patches into $K$ classes by maximizing the P-CLL term. However, the number of image patches varies greatly from class to class.
In order to equalize the number of similar patches in each class, we further adjust the classification.
If a class doesn't contain enough similar image patches, it will be merged with the nearest one.
If there are excessive patches in a class, k-means is used to divide the large class into some finer ones.
Instead of constructing low rank matrices, these preprocessed image patches in each class are used to record patches location information.
Finally, all the noisy image patches are clustered by the guide of the location information of the corresponding preprocessed ones.
This clustering process is somewhat complex, but the accuracy is undoubtedly higher. As a result, the rank of the created matrix by stacking patches from each class is lower.

Note: why not we use a denoising algorithm based on global patch matching?
Jin et al.~\cite{jin2017nonlocal,jin2018convergence} have proved that non-local means is a local method with a large neighborhood and there is an optimal size of search window.
Though global patch matching aims to acquire enough similar patches, the error rate of patch matching is intensely higher because of noise interference.
In the image preprocessing, if an image patch is classified into a wrong class,  the error will not be corrected in later  process.
For BM3D~\cite{dabov2007image}, the number of patches may  be insufficient and the result may not as good as global patch matching based methods (such as WNNM~\cite{gu2014weighted}, PCLR~\cite{chen2015external} and PGPD~\cite{Xu2016Patch}), 
but the error rate of patch matching is lower. 
The restoration result will be improved in the future denoising steps but the error will exist forever.
Another reason for not using global patch matching based methods is that they run slowly.

\subsection{Gaussian nuclear norm minimization for low rank modeling} 
\label{Sec: gnnm}
Consider the singular value decomposition (SVD) of a matrix $\mathbf{X}\in \mathbb{R}^{m\times n}$ of rank $r$:
$$
\mathbf{X} = \mathbf{U}\mathbf{\Lambda}(\mathbf{X})\mathbf{V}^{T},\,\, \mathbf{\Lambda}(\mathbf{X})=\diag(\{ \lambda_{\mathbf{X},i} \}_{1\leq i \leq r}),
$$
where $\mathbf{U}$ and $\mathbf{V}$ are respectively $m\times r$ and $n\times r$ matrices with orthonormal columns and the
singular values $\lambda_{\mathbf{X},i}$ are positive 
and satisfy that $\lambda_{\mathbf{X},1} \geq \lambda_{\mathbf{X},2} \geq \cdots \geq \lambda_{\mathbf{X},r}\geq 0  $.


Nuclear norm minimization (NNM)~\cite{fazel2003matrix} is an effective rank minimization.  The nuclear norm of a matrix $\mathbf{X}\in \mathbb{R}^{m\times n}$, denoted by $\|\mathbf{X}\|_{*}$, is defined as the sum of its singular values, i.e. 
\begin{equation}
   \| \mathbf{X}\|_{*}=\sum_{i=1}^{n}\lambda_{\mathbf{X},i}. 
\end{equation}
The NNM approach has been attracting significant attention due to its rapid development in both theory and implementation. The nuclear norm proximal (NNP) problem~\cite{cai2010singular} is
\begin{equation} \label{model nnp}
    \widetilde{\mathbf{X}} = \arg\min_{\mathbf{X}}\frac{1}{2}\|\mathbf{Y}-\mathbf{X}\|^{2}_{F}+ \mu\| \mathbf{X} \|_{*},
\end{equation}
where $\mu$ is a positive constant.

{\bf Theorem A } (Cai et al. \cite{cai2010singular}) For each $\mu>0$, the global optimum of model \eqref{model nnp} is 
\begin{equation}
\widetilde{\mathbf{X}}:= \mathbf{U}\widetilde{\bm{\Lambda}}(\mathbf{X})\mathbf{V}^{T},
\widetilde{\bm{\Lambda}}(\mathbf{X}) = \diag(\{(\lambda_{\mathbf{Y},i}-\mu)_{+}\}),
\end{equation} 
where $\lambda_{\mathbf{Y},i}$ denotes the $i$-th singular value of $\mathbf{Y}$ and $(\cdot)_{+} = \max(\cdot~,~0)$.
Albeit its success as aforementioned, NNM ignores the prior knowledge
we often have about singular values of a practical data matrix.
More specifically, larger singular values of an input data matrix quantify the information of its underlying principal
directions.
In order to improve the flexibility of NNM, Gu et al.~\cite{gu2014weighted} proposed the weighted nuclear norm. 
The weighted nuclear norm of a matrix $\mathbf{X}$ is defined as
\begin{equation}
    \| \mathbf{X} \|_{\mathbf{w},*} = \sum_{i} w_{i}\lambda_{\mathbf{X},i},
\end{equation}
where $\mathbf{w} = (w_{1},w_{2},\cdots,w_{r})$ 
is a non-negative weight assigned to $\lambda_{\mathbf{X},i}$.
The weight vector will enhance
the representation capability of the original nuclear norm.
The weighted nuclear norm proximal (WNNP) problem is given by
\begin{equation} \label{model wnnp}
    \widetilde{\mathbf{X}} = \arg\min_{\mathbf{X}}\frac{1}{2}\|\mathbf{Y}-\mathbf{X}\|^{2}_{F}+ \mu\| \mathbf{X} \|_{\mathbf{w},*}.
\end{equation}
{\bf Theorem B } (Gu et al. \cite{gu2014weighted}) If the weight $\mathbf{w}$ satisfies $0\leq w_{1}\leq w_{2}\leq \cdots \leq w_{r}  $, the global optimum of model \eqref{model wnnp} is 
\begin{equation}
\widetilde{\mathbf{X}}:= \mathbf{U}\widetilde{\bm{\Lambda}}(\mathbf{X})\mathbf{V}^{T},
\widetilde{\bm{\Lambda}}(\mathbf{X}) = \diag(\{(\lambda_{\mathbf{Y},i}-w_{i})_{+}\}).
\end{equation} 


The WNNM model is more difficult to optimize than conventional NNM ones due to the non-convexity of the involved weighted nuclear norm.

In this paper, we consider the proprieties of Gaussian noise. By tradition, the LRMA for Gaussian noise is to minimize $\|\mathbf{X}\|^{2}_{F}$. 
Note that $\|\mathbf{X}\|^{2}_{F}=\|\mathbf{X}\mathbf{X}^{T}\|_{*}$, we adapt NNM model  \eqref{model nnp} to Gaussian noise by utilizing $\mathbf{X}\mathbf{X}^{T}$ to take place of $\mathbf{X}$. Then we have Gaussian nuclear norm minimization  (GNNM):
\begin{equation}\label{Model Gaussian lr}
        \widetilde{\mathbf{X}} = \arg \min_{\mathbf{X}} \frac{1}{2}\|\mathbf{Y}\mathbf{Y}^{T}-\mathbf{X}\mathbf{X}^{T}\|_{F}^{2}
    +
    \mu\|\mathbf{X}\mathbf{X}^{T}\|_{*}.
\end{equation}
We prove that the Gaussian nuclear norm minimization problem can be equivalently transformed
to a quadratic programming (QP) problem with linear constraints. This allows us to easily reach the global optimum of the original problem by using off-the-shelf convex optimization solvers.
In order to solve the problem \eqref{Model Gaussian lr}, we  introduce the following theorem.

{\bf Theorem 1 } For each $\mu>0$, the global optimum of model \eqref{Model Gaussian lr} is 
\begin{equation}\label{modelsolution}
\widetilde{\mathbf{\mathbf{X}}}:= \mathbf{U}\widetilde{\bm{\Lambda}}(\mathbf{X})\mathbf{V}^{T},
\widetilde{\bm{\Lambda}}(\mathbf{X}) =  \diag\left(\left\{ \sqrt{(\lambda^{2}_{\mathbf{Y},i}-\mu)_{+}} \right\}\right).
\end{equation}

{\bf Proof } Considering  $\mathbf{X}\mathbf{X}^{T}$ as a new matrix and using Theorem A, we have that 
the global optimum of model \eqref{Model Gaussian lr} is 
$
\widetilde{\mathbf{\mathbf{X}\mathbf{X}}^{T}}:= \mathbf{U}\widetilde{\bm{\Lambda}}(\mathbf{X}\mathbf{X}^{T})\mathbf{U}^{T}$, $
\widetilde{\bm{\Lambda}}(\mathbf{X}\mathbf{X}^{T}) =  \diag\left(\left\{ {(\lambda^{2}_{\mathbf{Y},i}-\mu)_{+}} \right\}\right),$ where $\lambda^{2}_{\mathbf{Y},i}$ denotes the $i$-th singular value of $\mathbf{Y}\mathbf{Y}^{T}$. 
Let $\widetilde{\bm{\Lambda}}(\mathbf{X}) :=  \diag\left(\left\{ \sqrt{(\lambda^{2}_{\mathbf{Y},i}-\mu)_{+}} \right\}\right)$ and $\widetilde{\mathbf{\mathbf{X}}} := \mathbf{U} \widetilde{\bm{\Lambda}}(\mathbf{X}) \mathbf{V}^{T}$, it is easy to get
$\widetilde{\bm{\Lambda}}(\mathbf{X}\mathbf{X}^{T}) = \widetilde{\bm{\Lambda}}^{2}(\mathbf{X})$.
Then we have
\begin{eqnarray*}
  \widetilde{\mathbf{\mathbf{X}\mathbf{X}}^{T}}
  &=& \mathbf{U}\widetilde{\bm{\Lambda}}(\mathbf{X}\mathbf{X}^{T}) \mathbf{U}^{T}
  =
  \mathbf{U}\widetilde{\bm{\Lambda}}^{2}(\mathbf{X})\mathbf{U}^{T}
  \\&=& \mathbf{U}\widetilde{\bm{\Lambda}}(\mathbf{X}) \mathbf{V}^{T} \mathbf{V}\widetilde{\bm{\Lambda}}(\mathbf{X}) \mathbf{U}^{T}
  =
  \widetilde{\mathbf{\mathbf{X}}} \widetilde{\mathbf{\mathbf{X}}}^{T}.
\end{eqnarray*}
Therefore, $\widetilde{\mathbf{\mathbf{X}}} $ is also the  global optimum of model \eqref{Model Gaussian lr}.


A image contaminated by AGW noise i.e. model \eqref{Model AWG} can be recovered by the Theorem 1.
We next discuss the parameter $\mu$.
If the standard deviation of AWG noise is $\sigma$, we have
the shrinkage amount $\mu = m \sigma^2$. 
According to the AGW noise model \eqref{Model AWG}, we deduce the relationship between the eigenvalues of the covariance matrix for the noisy observation $\mathbf{Y}\mathbf{Y}^{T}$ and those for its latent clean image $\mathbf{X}\mathbf{X}^{T}$. 
Since $\bm{\varepsilon}$ is assumed to follow a Gaussian distribution with zero mean and variance $\sigma^2$,  the eigenvalue decomposition of the covariance matrix $\mathbf{Y}\mathbf{Y}^{T}$ can be expressed as
\begin{eqnarray*}
  \mathbf{U}\bm{\Lambda}(\mathbf{Y}\mathbf{Y}^{T})\mathbf{U}^{T} &=&  \mathbf{Y}\mathbf{Y}^{T} = (\mathbf{X} + \bm{\varepsilon})(\mathbf{X} + \bm{\varepsilon})^{T} \\
   &\approx&  \mathbf{X}\mathbf{X}^{T} +  \bm{\varepsilon} \bm{\varepsilon}^{T}
\\
&=&\mathbf{U}\bm{\Lambda}(\mathbf{X}\mathbf{X}^{T})\mathbf{U}^{T} + m\sigma^{n} \mathbf{I}\\
&=&\mathbf{U}(\bm{\Lambda}(\mathbf{X}\mathbf{X}^{T})+m\sigma^{2} \mathbf{I})\mathbf{U}^{T}. 
\end{eqnarray*}
 
Then we get the relationship between the eigenvalues of the noisy and clean images: 
\begin{equation}\label{eq lambda}
    \mathbf{\Lambda}^{2}(\mathbf{Y}) \approx \mathbf{\Lambda}^{2}(\mathbf{X}) + m\sigma^2\mathbf{I}.
    \end{equation}
Combining \eqref{modelsolution} and the equation above, we have $\mu = m\sigma^{2}$. As a result, model \eqref{Model Gaussian lr} can be rewritten as \begin{eqnarray} \label{Model Gaussian lrsigma}
    \widetilde{\mathbf{X}} =\arg \min_{\mathbf{X}} \frac{1}{2}\|\mathbf{Y}\mathbf{Y}^{T}-\mathbf{X}\mathbf{X}^{T}\|_{F}^{2} + m\sigma^2\|\mathbf{X}\mathbf{X}^{T}\|_{*}. 
\end{eqnarray}
    For any $\mathbf{X}_{k}$,  $k=1,2,\cdots, K$ in the model \eqref{Model ours}, we have that $\mu=q_{k}\sigma^2$.

\subsection{Optimization and algorithm}

We first divide all noise-free image patches into $K$ classes  and learn the parameter $\bm{\Theta}$ by GMM. Then there are only three unknowns in the proposed model \eqref{Model ours}: the latent  image $\mathbf{X}$, class label $\mathbf{c}$ and reshaped low rank matrices $\mathbf{X}_{k}$, $k=1,2,\cdots,K$. An alternating minimization procedure is used  to solve our model.
We start with the preprocessed image $\mathbf{X}^{pr}$.
In the $t$-th iteration, the solutions to the  alternating minimization scheme are
detailed as follows.

(1) Fix $\widetilde{\mathbf{X}}^{(t)}$,  solve for the class label $\widetilde{\mathbf{c}}^{(t)}$.

For each patch $(\mathbf{x}_{i}^{pr})^{(t)}$,
we use the probability density function of the learned
GMM to calculate the likelihood within each class,
\begin{eqnarray}
\begin{aligned}
p(k|(\mathbf{x}_{i}^{pr})^{(t)};\bm{\Theta}) = &\frac{{{\omega _k}{p_k}((\mathbf{x}_{i}^{pr})^{(t)}\left| {{\bm{\theta} _k})} \right.}}{{\sum\nolimits_j^K {{\omega _j}{p_j}((\mathbf{x}_{i}^{pr})^{(t)}\left| {{\bm{\theta} _j})} \right.} }}, \\
&k = 1, 2, \cdots ,K. 
\end{aligned}
\label{calculate c}
\end{eqnarray}
Then the $k$ with the maximum probability is assigned to $\widetilde{\mathbf{c}}^{(t)}$.

(2) Fix $\widetilde{\mathbf{c}}^{(t)}$,  we calculate low rank matrices $ \widetilde{\mathbf{X}}^{(t)}_{ {k}}$, $k=1,2,\cdots,K$, according to
\begin{eqnarray}
\begin{aligned}
{  \widetilde{\mathbf{X}}^{(t)}_{ {k}}} = \arg\mathop {\min }\limits_{ \mathbf{X}_{ {k}}}  \Big(& \left\|  \mathbf{Y}^{(t)}_{k}(\mathbf{Y}_{k} ^{(t)})^{T}- \mathbf{X}_{k}\mathbf{X}_{k}^{T} \right\|_F^2\\
&+ q_{k}(\sigma^{(t)})^2\left\| \mathbf{X}_{k}\mathbf{X}_{k}^{T} \right\|_ * \Big).
\end{aligned}
\end{eqnarray}
Like Gu et al.~\cite{gu2014weighted}, we solve the low rank denoising model \eqref{Model ours} by the formulas as follows:
\begin{eqnarray} \label{calculate X}
&&\widetilde{\mathbf{X}}_{k}^{(t)}= \mathbf{U}\widetilde{\mathbf{\Lambda}}(\mathbf{X}_{k}^{(t)})\mathbf{V}^{T},\\
 &&\widetilde{\mathbf{\Lambda}}(\mathbf{X}_{k}^{(t)})=\sqrt{\max\left(({\lambda}_{\mathbf{Y}^{(t)}_{k},i})^2-q_{k} (\sigma^{(t)})^2, 0\right)}, \nonumber
\end{eqnarray}
here  ${\mathbf{\Lambda}}(\mathbf{Y}_{k}^{(t)})=\diag (\{\lambda_{\mathbf{Y}_{k}^{(t),i}}\})$ is the singular value matrix of $\mathbf{Y}^{(t)}_k$, $\mathbf{U}{\mathbf{\Lambda}}(\mathbf{Y}_{k}^{(t)}) \mathbf{V}^T$ denotes the singular value decomposition (SVD) of $\mathbf{Y}^{(t)}_k$, $q_{k}$ is the number of similar patches, 
and $(\sigma^{(t)})^2$ is the variance of image $\mathbf{Y}^{(t)}$ in the $t$-th iteration.

(3) Fix $\widetilde{\mathbf{X}}^{(t)}_{k}$, $k=1,2,\cdots,K$, a 
weighted averaging method is used to reconstruct the  estimated image $\widetilde{\mathbf{X}}^{(t)}$ by aggregating all denoised patches.
The value of the $i$-th pixel $ \widetilde{x}^{(t)}_{i}$ in the reference image $\widetilde{\mathbf{X}}^{(t)}$ is calculated as a weighted average as follows,
\begin{eqnarray}
    \widetilde{x}^{(t)}_{i} = \left(\sum\limits_{k=1}^{K} \sum\limits_{j=1}^{q_{k}} \mathbf{W}_{kj} \widetilde{\mathbf{X}}^{(t)}_{kji}\right) /  \left(\sum\limits_{k=1}^{K} \sum\limits_{j=1}^{q_{k}} \mathbf{W}_{kj}\right),
    \label{calculate image}
\end{eqnarray}
where $\widetilde{\mathbf{X}}^{(t)}_{kji}$
refers to the denoised image intensity of the $i$-th pixel in the $j$-th patch of $k$-th patch group , and the summation is carried out over all patches that overlap with the $i$-th pixel.
The empirical weight $\mathbf{W}$ is given as
\begin{eqnarray}
    \mathbf{W}_{kj} = \begin{cases}
1-s_k / {q_{k}},    &s_k <  q_{k};\\
1  / {q_{k}}                    ,   &s_k = q_{k},
\end{cases}
\end{eqnarray}
$s_{k}$ is the rank of matrix $\widetilde{\mathbf{X}}^{(t)}_{k}$, $ q_{k}$ is the number of similar patches in the $k$-th class.

In addition, at the beginning of each iteration $t$, the desired denoised image, preprocessed image and standard deviation are updated by
\begin{eqnarray}
   \mathbf{Y}^{(t)}
   =\widetilde{\mathbf{X}}^{(t-1)}+\alpha (\mathbf{Y}-\widetilde{\mathbf{X}}^{(t-1)}),
    \label{update Y}
\end{eqnarray}
\begin{eqnarray}
(\mathbf{X}^{pr})^{(t)}=(\mathbf{X}^{pr})^{(t-1)}+\alpha (\mathbf{Y}-(\mathbf{X}^{pr})^{(t-1)}),
\label{update Xpr}
\end{eqnarray}
\begin{eqnarray}
\sigma^{(t)}
=
\beta \sqrt{\sigma^2-\|\mathbf{Y}-\mathbf{Y}^{(t)}\|_{F}^2},
\label{update sigma}
\end{eqnarray}
where $ t $ means  the $t$-th iteration, $ \alpha $ is the iteration regularization parameter, $ \mathbf{Y} $ is the noisy image, $\widetilde{\mathbf{X}}^{(t)} $ is the denoised image in the $t$-th iteration, 
 $\sigma^2$ is the noise variance of the input noisy image, 
and  $ \beta $ is the scale factor of controlling the re-estimation noise standard deviation $\sigma^{(t)}$. The complete optimization process is shown in Algorithm \ref{algorithm a}.

\begin{algorithm}[t]
\label{algorithm a}
\caption{ \small{Preprocessed image based GMM prior low rank regularization (PG-LR)} }
\KwIn{Noisy image $ \mathbf{Y}$, $\sigma$, $\bm{\Theta}$, \\
\quad Intialize: 1)~$\alpha, \beta , MaxIter, K$ \\
\quad \quad \quad \quad \quad \quad and~$\sigma^{(0)} = \sigma , ~\widetilde{\mathbf{X}}^{(0)} =  \mathbf{Y}, ~ \mathbf{Y}^{(0)} =  \mathbf{Y}  $\\
\quad \quad \quad \quad \quad   2) BM3D preprocessed image~$(\mathbf{X}^{pr})^{(0)}$
}

 \For {~$t = 1: MaxIter$ }
 {
Interative regularization $\mathbf{Y}^{(t)} $ and $ (\mathbf{X}^{pr})^{(t)}$ by Eqs. \eqref{update Y}  and \eqref{update Xpr}, respectively

 \For {~$k = 1:K$  }
{
Update~$\mathbf{\Sigma}^{(t)}_k = \mathbf{\Sigma}_k + {(\sigma^{(t-1)}) ^2}\mathbf{I}$

Calculate~$p(k|(\mathbf{x}_{i}^{pr})^{(t)};\bm{\Theta})$ via Eq.\eqref{calculate c}}  
Compute $\widetilde{\mathbf{c}}^{(t)}$ by finding the class that has the largest conditional probability

 \For {~$k = 1:K$  }
{

Create $ {\mathbf{Y}}^{(t)}_k$ by stacking patches 
from the $k$-th class

Sigular value decomposition \\
 \quad \quad  $[\mathbf{U},\mathbf{\Lambda}(\mathbf{Y}_{k}^{(t)}),\mathbf{V}]
=\mathrm{SVD}({\mathbf{Y}}^{(t)}_k)$

Get the estimation~$\widetilde{\mathbf{X}}^{(t)}_k=\mathbf{U}\widetilde{\mathbf{\Lambda}}(\mathbf{X}_{k}^{(t)})\mathbf{V}^{T}$ \\~~ via Eq.  \eqref{calculate X}
}

Aggregate~$\widetilde{\mathbf {X}}^{(t)}_k$ to form denoised image~$\widetilde{\mathbf{X}}^{(t)}$  via Eq. \eqref{calculate image}

Update $\sigma^{(t)}$ by Eq. \eqref{update sigma}
}
\KwOut{Restored image~$ \widetilde{\mathbf{X}}$}
\end{algorithm}

\subsection{Complexity and convergence}
Suppose that there are $m \times n$ patches extracted from the whole image and $s\times s$ pixels in each patch. The main computational cost in each iteration is mainly composed of three parts. Firstly, classifying patches into $K$ classes costs $O(mnKs^6)$, where the covariance matrix determinant needs $O(s^6)$ calculations. Next, the computation of the low rank matrix reconstruction by SVD for $K$ clusters takes $O(mnKs^4)$. Finally, averaging all the grouping similar patches needs $O(mns^2)$ computations. Thus the computation complexity  is $O(mnKs^6)  +O(mnKs^4)+O(
mns^2)$ and the classification step takes most computations of the algorithm.

\begin{figure}
    \begin{center}
    \begin{tabular}{c@{\hskip 2pt}c@{\hskip 2pt}c@{\hskip 2pt}c@{\hskip 2pt}c@{\hskip 2pt}c@{\hskip 2pt}c@{\hskip 2pt}c@{\hskip 2pt}c@{\hskip 2pt}c}
\includegraphics[width=0.24\textwidth]{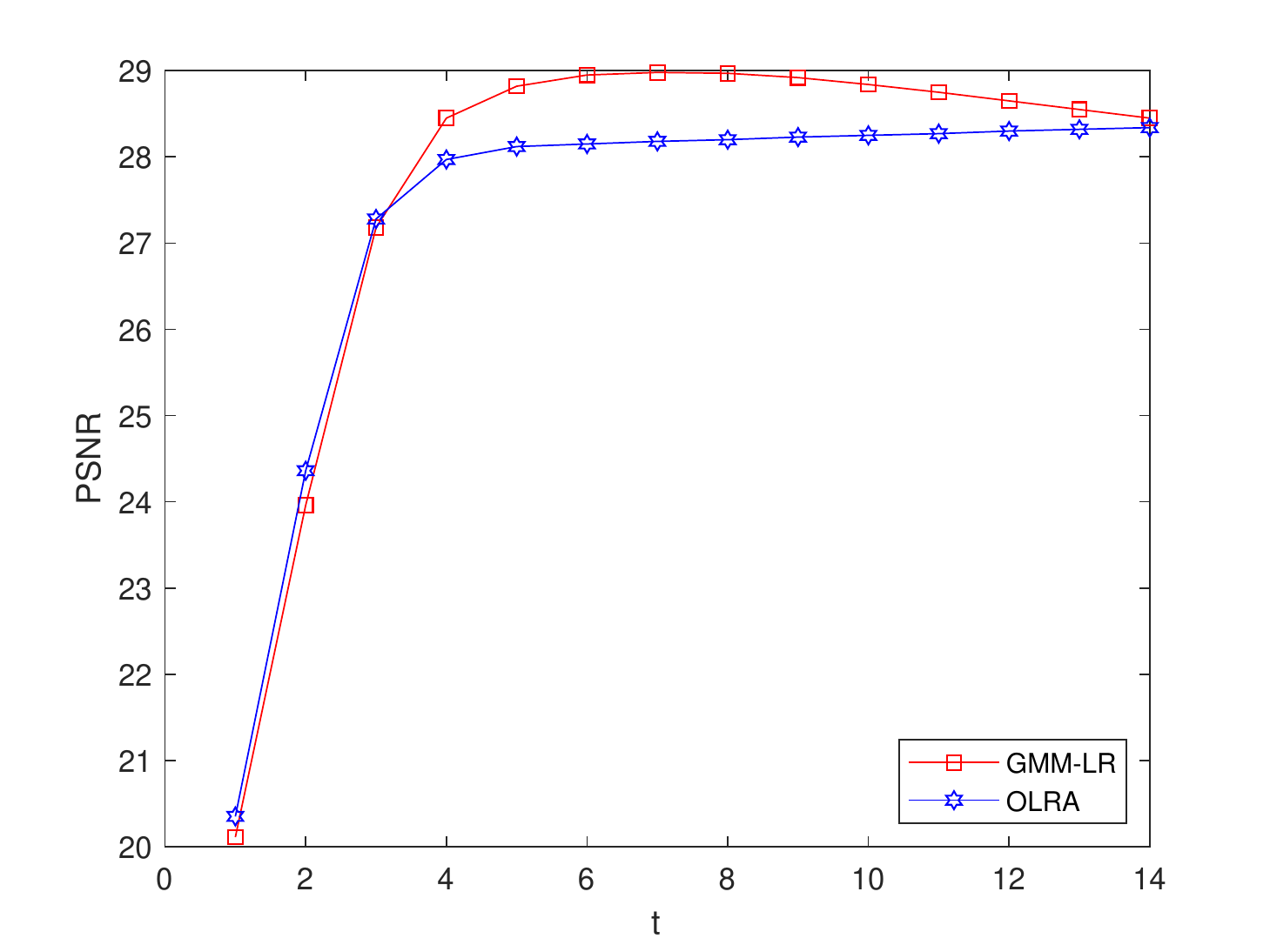}&
\includegraphics[width=0.24\textwidth]{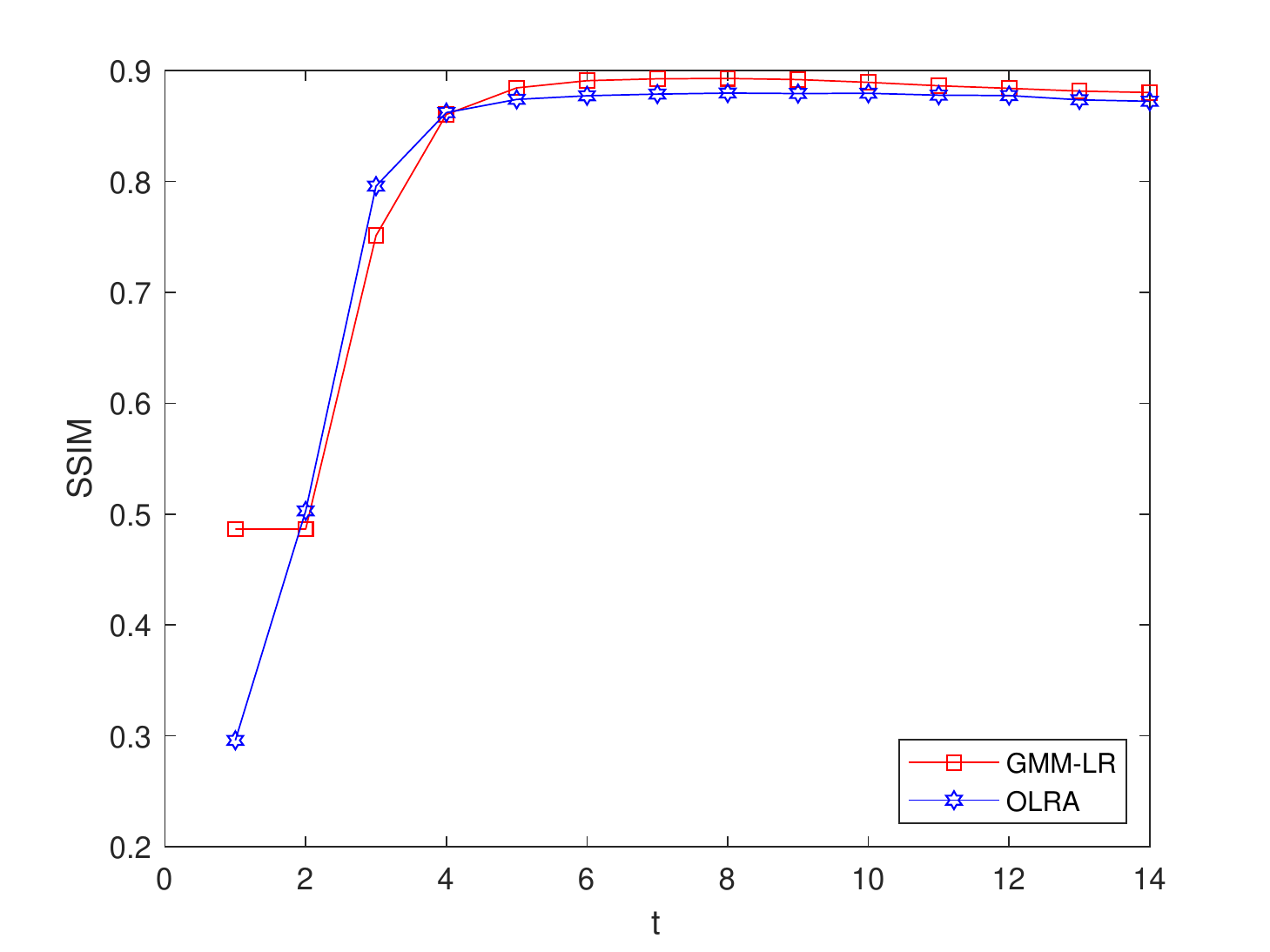}\vspace{-5pt}\\
\scriptsize (a) &
\scriptsize (b) \vspace{-5pt}&
 \end{tabular}
 \end{center}
 \caption{  \small{(a) and (b)  are graphs of  the PSNR  and SSIM variation curve with iteration $t\in[1,  14]$, $\alpha=0.10$,~$\beta=0.62$.
 } }
\label{Fig parameter mu}
\end{figure}

Experience shows our method has a rapid convergence and reaches the optimal solution  at the $5$-th or $6$-th iteration although it is non-convex. Figure \ref{Fig parameter mu} compares the convergence speed between our algorithm and OLRA~\cite{zhang2018kernel}. In the comparative experiment, the Cameraman image with intricate details was selected. Figure \ref{Fig parameter mu} (a) and (b) describe the variation curve of PSNR and SSIM values, respectively. It's obvious that our method is always superior than OLRA~\cite{zhang2018kernel} after the $5$-th iteration.
As shown in Table \ref{tab:table z}, our algorithm is almost the fastest among all low rank based methods.

\begin{table}[t] \scriptsize 
\centering
\caption{ \small{The average computation time(s) for restoring grayscale image with noise level $\sigma=50$ on a desktop (Intel(R) Core(TM)
i5-8250 CPU @1.60 GHz) with MATLAB 2019b. } }\label{tab:table z}
\renewcommand{\tablename}{}
 \begin{tabular}{lccccccc}
 \toprule
Image size &$256\times256$ &  $512\times512$ \\
 \midrule
 BM3D\cite{dabov2007image}& 1.26  & 3.22\\
WNNM~\cite{gu2014weighted} &  117.97 & 677.67 \\
 PCLR~\cite{chen2015external} & 196.11&1090.40\\
  OLRA ~\cite{zhang2018kernel} &  202.20 &  1014.63\\
  DnCNN~\cite{zhang2017beyond} &  18.83 &  15.93\\
  FFDNet~\cite{Zhang2018FFDNet} & 16.49 & 17.23\\
   Ours& 129.45 &  614.10\\
\bottomrule
\end{tabular}
\centering
\end{table}

\begin{figure}
    \centering
    \begin{tabular}{c@{\hskip 1pt}c@{\hskip 1pt}c@{\hskip 1pt}c@{\hskip 1pt}c@{\hskip 1pt}c@{\hskip 1pt}c@{\hskip 1pt}c@{\hskip 1pt}c@{\hskip 1pt}c@{\hskip 1pt}c@{\hskip 1pt}c@{\hskip 1pt}c@{\hskip 1pt}c@{\hskip 1pt}c@{\hskip 1pt}c@{\hskip 1pt}c@{\hskip 1pt}c@{\hskip 1pt}c@{\hskip 1pt}c@{\hskip 1pt}c@{\hskip 1pt}c@{\hskip 1pt}c@{\hskip 1pt}c@{\hskip 1pt}c}
\includegraphics[width=0.045\textwidth]{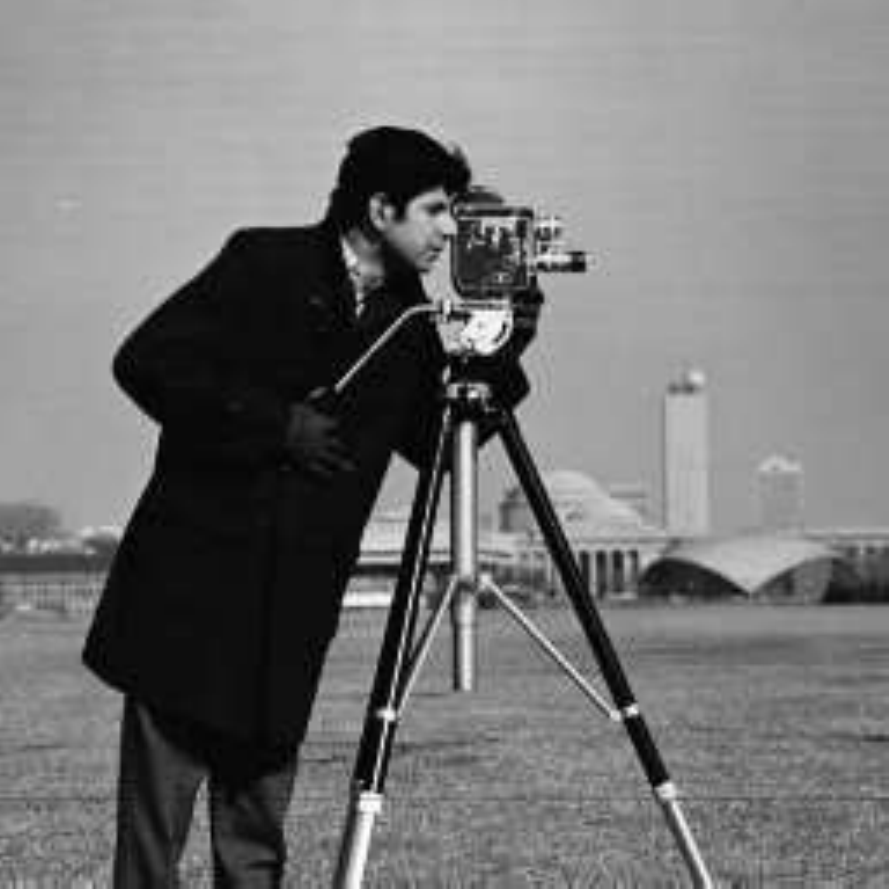}&
\includegraphics[width=0.045\textwidth]{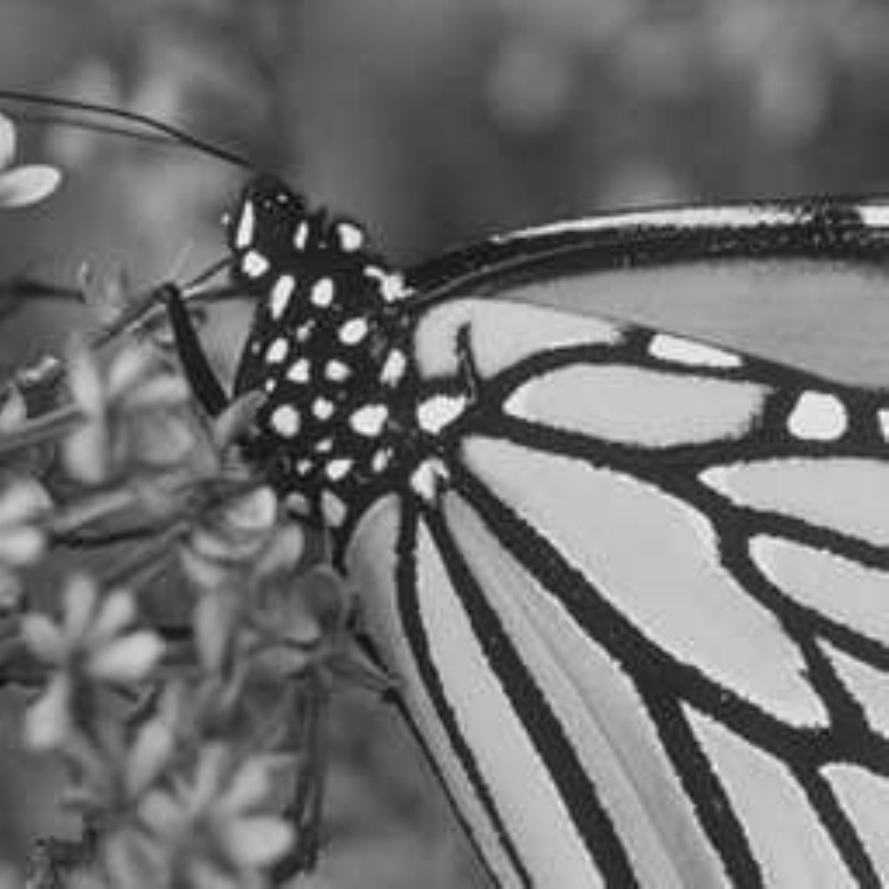}&
\includegraphics[width=0.045\textwidth]{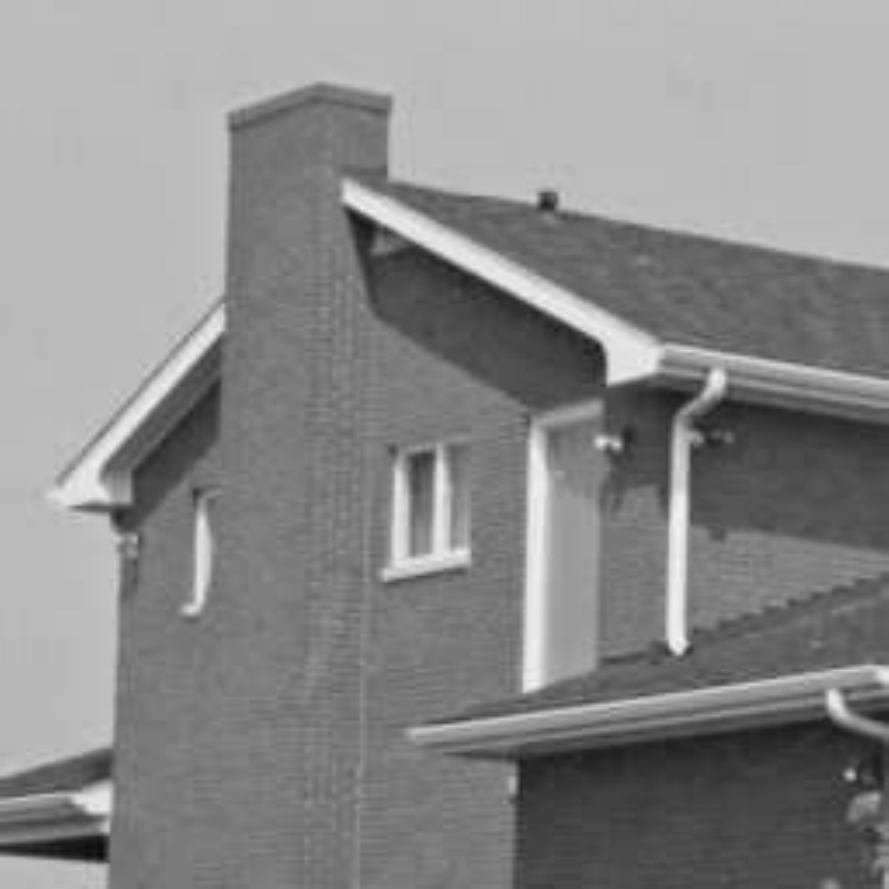}&
\includegraphics[width=0.045\textwidth]{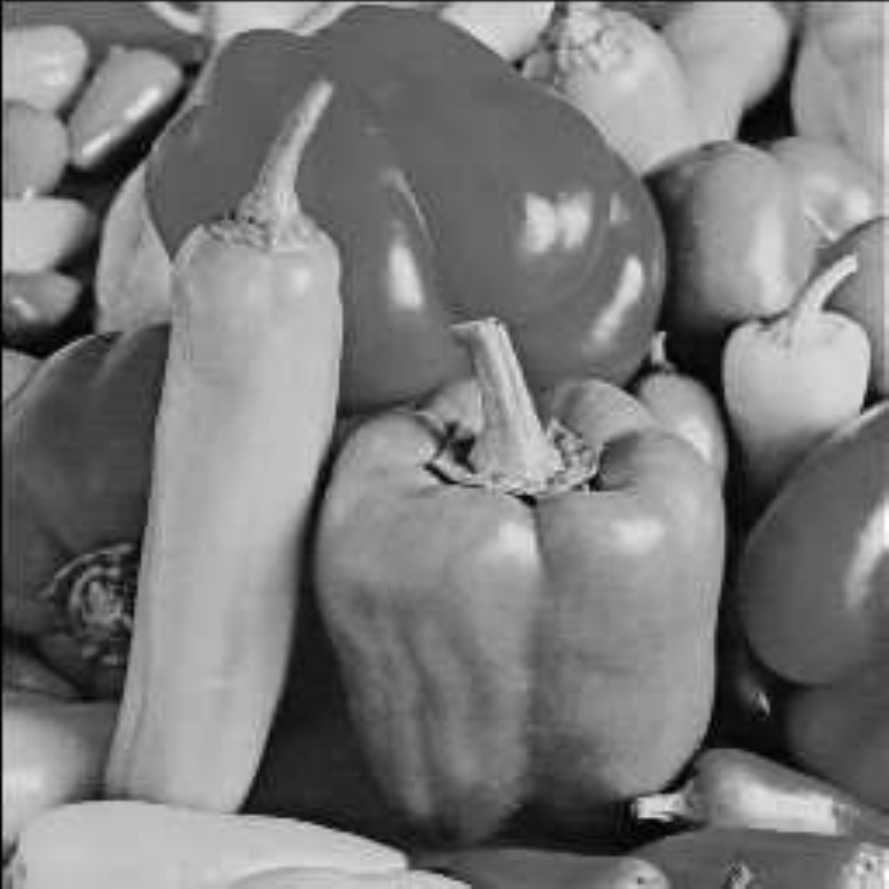}&
\includegraphics[width=0.045\textwidth]{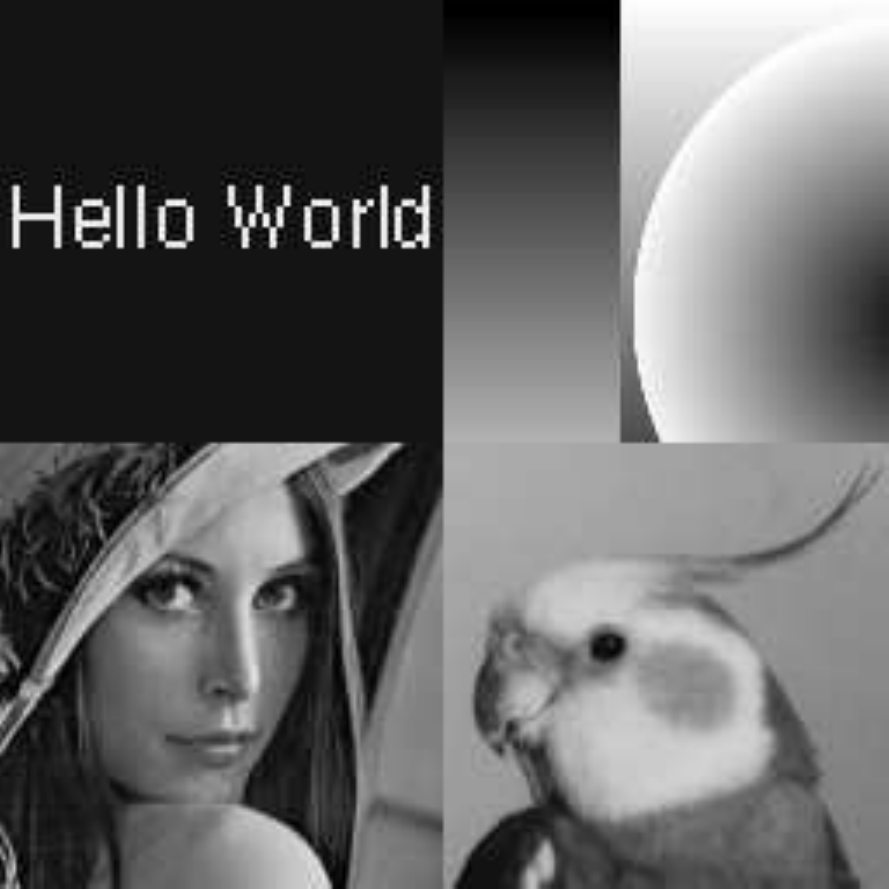}&
\includegraphics[width=0.045\textwidth]{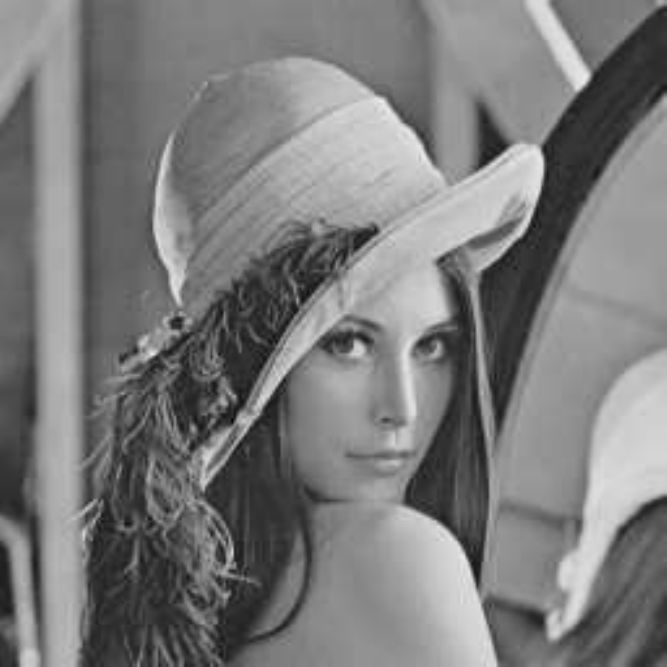}&
\includegraphics[width=0.045\textwidth]{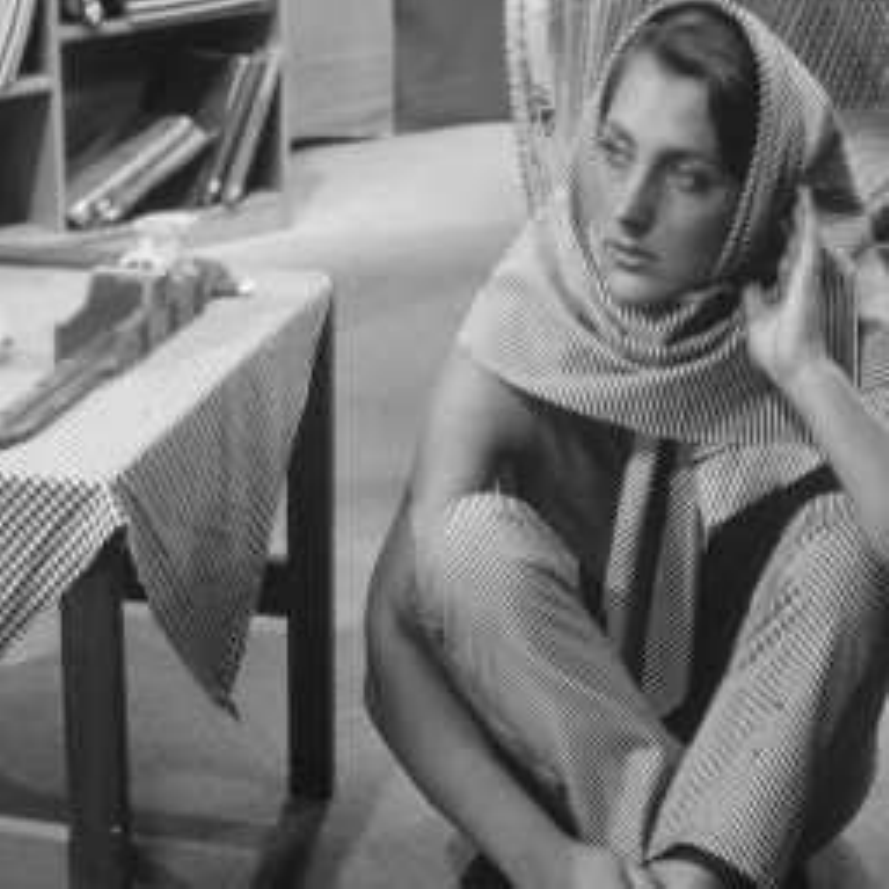}&
\includegraphics[width=0.045\textwidth]{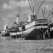}&
\includegraphics[width=0.045\textwidth]{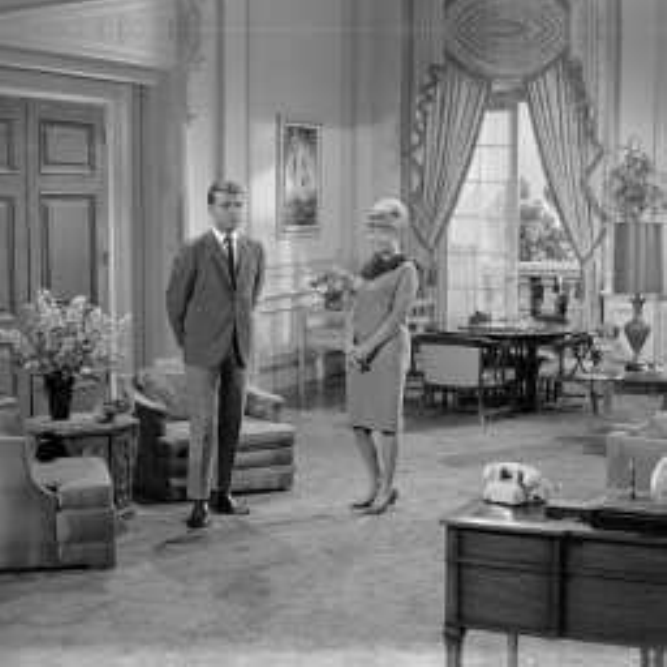}&
\includegraphics[width=0.045\textwidth]{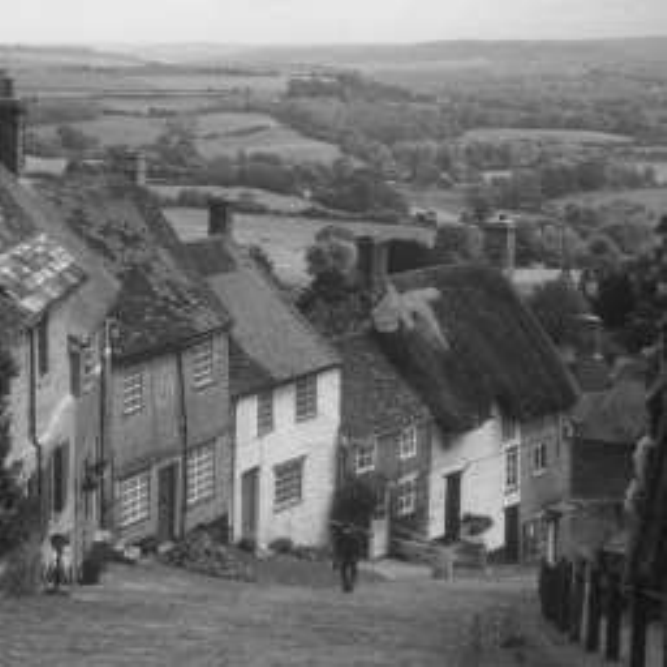}& \vspace{-5pt}\\
\tiny Cameraman & \tiny Monarch &\tiny House &\tiny Peppers &\tiny Montage&
\tiny Lena &\tiny Barbara &\tiny Boat &\tiny Couple &\tiny Hill
    \end{tabular}
    \caption{  \small{ All test images of the simulation experiment. } }
    \label{Fig testimages}
\end{figure}

\begin{table*}[t] \scriptsize
\centering
\caption{ \small
Comparison in  PSNR$/$SSIM values  between the state-of-the-art denoising method and ours with noise level $\sigma=25, 50, 75, 100$.
The best PSNR$/$SSIM values in all traditional methods are marked in \textbf{bold} and the best result of each row  is marked in {\color{red}red}. A transitional method wins in the raw marked in both {
\textbf{bold}} and  {\color{red}\textbf{red}}. As references we also include the results of DnCNN~\cite{zhang2017beyond} 
and FFDNet~\cite{Zhang2018FFDNet}. 
 {\bf Notes:} 1.  DnCNN* \cite{zhang2017beyond} is the result of directly using the DnCNN network in the matlab 2019b deep learning toolbox; 2.Results limited to $\sigma\le75$, as the DnCNN network is not trained beyond that level.}
\label{tab:table a}
\begin{tabular}{|c|c|c|c|c|c|c|c|c|c|}
\hline
&&\multicolumn{5}{c|}{
\textbf{ Traditional algorithm }
} &\multicolumn{3}{c|}{ \textbf{ Deep Learning }}  \\
\hline \hline
$\sigma$ & PSNR/SSIM&BM3D~\cite{dabov2007image}&WNNM~\cite{gu2014weighted}&PCLR~\cite{chen2015external} &OLRA~\cite{zhang2018kernel}&Ours&FFDNet~\cite{Zhang2018FFDNet}&DnCNN~\cite{zhang2017beyond}&DnCNN*\cite{zhang2017beyond}\\
\hline
\multirow{11}{*}{25}
&Cameraman &29.45/0.854&29.63/0.860&29.66/0.863&29.58/0.863&\textbf{29.90}/\textbf{0.870}&30.05/{\color{red}0.877}&  {\color{red} 30.18}/0.876&28.87/0.823\\
\cline{2-10}
&Monarch &29.25/0.890&29.85/0.909&29.75/0.908&29.92/0.912&{\color{red}\textbf{30.26}}/{\color{red}\textbf{0.918}}&30.14/0.917&30.23/0.917&29.26/0.879 \\
\cline{2-10}
&House &32.86/0.859&33.23/0.861&32.99/0.859&33.25/0.863&{\color{red}\textbf{33.50}}/{\color{red}\textbf{0.864}}&33.27/0.862&33.08/0.860&31.65/0.825\\
\cline{2-10}
&Peppers &30.16/0.868&30.40/0.873&30.46/0.875&30.35/0.874&\textbf{30.76}/\textbf{0.881}&30.79/{\color{red}0.884}&{\color{red}30.84}/0.881&29.49/0.840\\
\cline{2-10}
&Montage &32.37/0.926&\textbf{32.74}/0.933&32.36/0.930&32.26/0.929&\textbf{32.74}/\textbf{0.934} &32.84/{\color{red}0.941}&{\color{red}32.95}/0.937&30.68/0.879\\
\cline{2-10}
&Lena &32.08/0.861&32.25/0.866&32.17/0.863&32.30/0.867&\textbf{32.55}/\textbf{0.868} &{\color{red}32.59}/{\color{red}0.874}&32.44/0.869&31.12/0.824\\
\cline{2-10}
&Barbara &30.72/0.874&{\color{red}\textbf{31.24}}/{\color{red}\textbf{0.898}}&30.65/0.887&31.20/0.897&31.07/0.894&29.98/0.879&30.00/0.878&26.78/0.785\\
\cline{2-10}
&Boat &29.91/0.801&29.83/0.795&30.00/0.800&29.95/0.803&{\color{red}\textbf{30.34}}/\textbf{0.811} &30.23/{\color{red}0.812}&30.21/0.809&29.45/0.780\\
\cline{2-10}
&Couple &29.72/0.820&29.82/0.820&29.73/0.814&29.77/0.819&\textbf{30.11}/\textbf{0.826} &{\color{red}30.18}/{\color{red}0.835}&30.12/0.830&29.08/0.788\\
\cline{2-10}
&Hill &29.85/0.775&29.95/0.776&29.84/0.771&29.94/0.778&{\color{red}\textbf{30.24}}/{\color{red}\textbf{0.785}}&30.04/0.782&  30.03/0.780&29.43/0.758\\
\cline{2-10}
&Average &30.64/0.855&30.91/0.860&30.76/0.857&30.85/0.861&{\color{red}\textbf{31.15}}/\textbf{0.865}&31.01/{\color{red}0.866}&  31.01/0.864&29.58/0.818\\
\hline
\hline
\multirow{11}{*}{50}
&Cameraman &26.12/0.782&26.41/0.785&26.55/\textbf{0.795}&26.35/0.780&\textbf{26.83}/0.794&{\color{red}27.02}/{\color{red}0.811}&27.00/0.805&25.75/0.704\\
\cline{2-10}
&Monarch &25.82/0.820&26.32/0.835&26.25/0.837&26.21/0.835&\textbf{26.67}/\textbf{0.849} &{\color{red}26.88}/{\color{red}0.854}&26.76/0.846&25.46/0.767\\
\cline{2-10}
&House &29.69/0.812&30.33/\textbf{0.823}&29.77/0.819&30.19/0.816&\textbf{30.40}/0.818 &{\color{red}30.43}/{\color{red}0.829}&30.01/0.820&27.98/0.711\\
\cline{2-10}
&Peppers &26.68/0.794&26.91/0.801&27.03/0.808&27.06/0.800&{\color{red}\textbf{27.69}}/{\color{red}\textbf{0.820}}&27.43/{\color{red}0.820}&27.29/0.810&25.93/0.724\\
\cline{2-10}
&Montage &27.90/0.861&28.27/0.875&28.20/0.877&28.34/0.872&\textbf{28.95}/\textbf{0.883}&{\color{red}29.18}/{\color{red}0.898}&28.95/0.882&26.85/0.772\\
\cline{2-10}
&Lena &29.05/0.799&29.25/0.806&29.12/0.801&29.01/0.798&{\color{red}\textbf{29.68}}/\textbf{0.811} &{\color{red}29.68}/{\color{red}0.823}&29.36/0.811&27.78/0.705\\
\cline{2-10}
&Barbara &27.23/0.795&27.79/0.820&27.11/0.798&27.65/0.811&{\color{red}\textbf{27.90}}/{\color{red}\textbf{0.821}}&26.48/0.779&  26.23/0.770&24.08/0.642\\
\cline{2-10}
&Boat &26.78/0.705&26.97/0.708&26.99/0.708&26.84/0.701&{\color{red}\textbf{27.47}}/\textbf{0.721}&27.32/{\color{red}0.726}&  27.19/0.718&26.30/0.660\\
\cline{2-10}
&Couple &26.46/0.707&26.64/0.714&26.55/0.706&26.65/0.714&{\color{red}\textbf{27.16}}/\textbf{0.727}&27.07{\color{red}/0.737}&  26.89/0.724&25.83/0.651\\
\cline{2-10}
&Hill &27.19/0.675&27.33/0.676&27.23/0.672&27.24/0.671&{\color{red}\textbf{27.85}}/{\color{red}\textbf{0.690}}&27.55/0.688&  27.44/0.683&26.55/0.631\\
\cline{2-10}
&Average &27.29/0.775&27.62/0.784&27.48/0.782&27.55/0.780&{\color{red}\textbf{28.06}}/\textbf{0.793}&27.90/{\color{red}0.796}& 27.71/0.787&26.25/0.697\\
\hline
\hline
\multirow{11}{*}{75}
&Cameraman &24.33/0.734&24.55/0.735&24.74/0.747&24.65/0.741&\textbf{25.16}/\textbf{0.753}&{\color{red}25.29}/{\color{red}0.770}&  25.09/0.759&22.48/0.466\\
\cline{2-10}
&Monarch &23.91/0.756&24.31/0.775&24.28/0.781&24.10/0.776&\textbf{24.71}/\textbf{0.796}&{\color{red}24.99}/{\color{red}0.801}&  24.70/0.783&22.34/0.588\\
\cline{2-10}
&House &27.51/0.765&28.24/0.789&27.57/0.785&28.38/0.795&{\color{red}\textbf{28.78}}/{\color{red}\textbf{0.803}}&28.43/0.796&  27.83/0.776&23.82/0.466\\
\cline{2-10}
&Peppers &24.73/0.737&24.92/0.742&25.06/0.762&24.95/0.748&{\color{red}\textbf{25.67}}/\textbf{0.768}&25.39/{\color{red}0.769}&  25.15/0.752&22.83/0.526\\
\cline{2-10}
&Montage &25.52/0.800&25.72/0.821&25.70/0.827&25.56/0.831&{\textbf{26.57}}/\textbf{0.851}&{\color{red}26.84}/{\color{red}0.857}&  26.30/0.829&23.36/0.516\\
\cline{2-10}
&Lena &27.26/0.752&27.54/0.766&27.41/0.766&25.96/0.766&{\color{red}\textbf{28.16}}/\textbf{0.776}&  27.57/0.767&27.97/{\color{red}0.786}&24.03/0.471\\
\cline{2-10}
&Barbara &25.12/0.711&25.81/0.749&25.07/0.722&25.68/0.740&{\color{red}\textbf{26.03}}/{\color{red}\textbf{0.753}}&24.24/0.685&  23.62/0.654&21.95/0.462\\
\cline{2-10}
&Boat &25.12/0.641&25.30/0.647&25.29/0.649&25.00/0.634&{\color{red}\textbf{25.78}}/\textbf{0.658}&25.64/{\color{red}0.666}&  25.47/0.655&23.01/0.461\\
\cline{2-10}
&Couple &24.70/0.626&24.86/0.637&24.80/0.629&24.71/0.627&{\color{red}\textbf{25.43}}/\textbf{0.651}&25.29/{\color{red}0.662}&  24.99/0.640&22.87/0.466\\
\cline{2-10}
&Hill &25.68/0.612&25.88/0.619&25.82/0.616&25.59/0.604&{\color{red}\textbf{26.36}}/\textbf{0.625}&26.15/{\color{red}0.633}&  25.95/0.621&23.59/0.454\\
\cline{2-10}
&Average &25.39/0.713&25.71/0.728&25.57/0.728&25.46/0.726&{\color{red}\textbf{26.27}}/{\color{red}\textbf{0.743}}&26.02/0.742&  25.67/0.724&23.03/0.487\\
\hline
\hline
\multirow{11}{*}{100}
&Cameraman &23.07/0.692&23.36/0.697&23.48/0.715&23.32/0.700& {\color{red}\textbf{23.96}}/\textbf{0.721}&{\color{red}23.96}/{\color{red}0.740}&  -- &  -- \\
\cline{2-10}
&Monarch &22.52/0.702&22.95/0.726&22.93/0.737&22.86/0.727&{\color{red}\textbf{23.62}}/\textbf{0.758}&23.61/{\color{red}0.759}&  --&  --\\
\cline{2-10}
&House &25.87/0.720&26.66/0.754&25.97/0.750&26.66/0.755&{\color{red}\textbf{27.22}}/{\color{red}\textbf{0.770}}&26.91/0.768 &  --&  -- \\
\cline{2-10}
&Peppers &23.39/0.688&23.45/0.698&23.70/0.725&23.66/0.711&{\color{red}\textbf{24.45}}/{\color{red}\textbf{0.743}}&23.96/0.733&  --&  -- \\
\cline{2-10}
&Montage &23.89/0.747&24.16/0.777&24.23/0.794&23.86/0.783&\textbf{25.07}/\textbf{0.818}&{\color{red}25.12}/{\color{red}0.824}&  -- &  --\\
\cline{2-10}
&Lena &25.95/0.709&26.21/0.726&26.16/0.736&26.96/0.725&{\color{red}\textbf{27.02}}/\textbf{0.741}&26.77/{\color{red}0.756} &  --&  --\\
\cline{2-10}
&Barbara &23.62/0.643&24.37/0.686&23.65/0.660&24.10/0.672&{\color{red}\textbf{24.63}}/{\color{red}\textbf{0.691}}&22.82/0.614&  --&  -- \\
\cline{2-10}
&Boat &23.97/0.594&24.11/0.598&24.16/0.607&23.83/0.594&{\color{red}\textbf{24.77}}/\textbf{0.618}&24.53/{\color{red}0.624}&  --&  -- \\
\cline{2-10}
&Couple &23.51/0.567&23.56/0.570&23.65/0.575&23.55/0.567&{\color{red}\textbf{24.37}}/\textbf{0.597}&24.05/{\color{red}0.600}&  --&  --\\
\cline{2-10}
&Hill &24.58/0.565&24.76/0.572&24.83/0.576&24.53/0.562&{\color{red}\textbf{25.45}}/\textbf{0.587}&25.15/{\color{red}0.591}&  -- &  --\\
\cline{2-10}
&Average &24.04/0.663&24.36/0.680&24.28/0.688&24.23/0.680&{\color{red}\textbf{25.06}}/{\color{red}\textbf{0.704}}&24.69/0.701&  -- &  --\\

\hline
\end{tabular}
\centering
\end{table*}

\begin{figure*}
    \centering
    \begin{tabular}{c@{\hskip 1pt}c@{\hskip 1pt}c@{\hskip 1pt}c@{\hskip 1pt}c@{\hskip 1pt}c@{\hskip 1pt}c@{\hskip 1pt}c@{\hskip 1pt}c@{\hskip 1pt}c}
\includegraphics[width=0.170\textwidth]{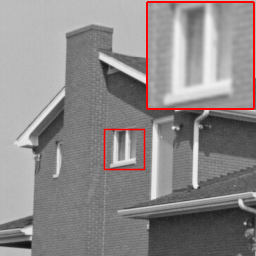}&
\includegraphics[width=0.170\textwidth]{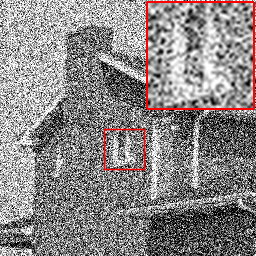}&
\includegraphics[width=0.170\textwidth]{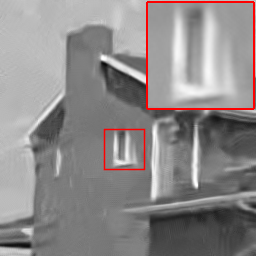}&
\includegraphics[width=0.170\textwidth]{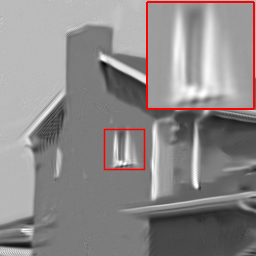}&
\includegraphics[width=0.170\textwidth]{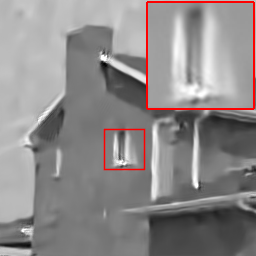}\vspace{-5pt}\\
\scriptsize (a) Original &
\scriptsize (b) Noisy image &
\scriptsize (c) BM3D\cite{dabov2007image}~\scriptsize{27.51/0.765}&
\scriptsize (d) WNNM\cite{gu2014weighted}~\scriptsize{28.24/0.789}&
\scriptsize (e) PCLR\cite{chen2015external}~\scriptsize{27.57/0.785}\\
\includegraphics[width=0.170\textwidth]{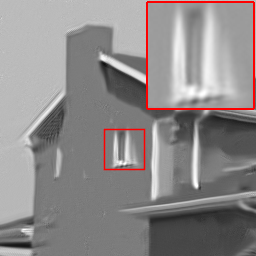}&
\includegraphics[width=0.170\textwidth]{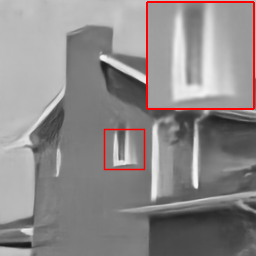}&
\includegraphics[width=0.170\textwidth]{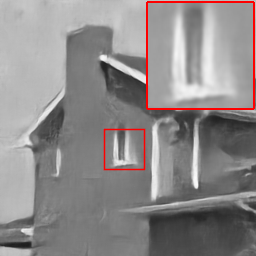}&
\includegraphics[width=0.170\textwidth]{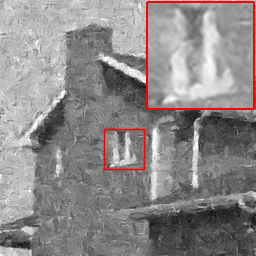}&
\includegraphics[width=0.170\textwidth]{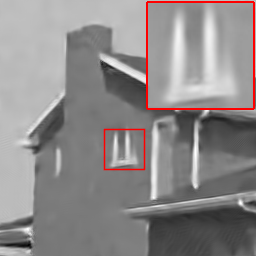}\vspace{-5pt}\\
\scriptsize (f) OLRA\cite{zhang2018kernel}~\scriptsize{28.38/0.795}&
\scriptsize (g) FFDNet\cite{Zhang2018FFDNet}~\scriptsize{28.43/0.796}&
\scriptsize (h) DnCNN\cite{zhang2017beyond}~\scriptsize{27.83/0.776}&
\scriptsize (i) DnCNN*\cite{zhang2017beyond}~\scriptsize{23.82/0.466}&
\scriptsize (j) Ours~\scriptsize{28.78/0.803}&\\
 \end{tabular}
 \caption{  \small{{Denoising results on image House by different methods with noise level $\sigma  = 75$. } }
}
\label{Fig: house}
\end{figure*}

\begin{figure*}
    \centering
    \begin{tabular}{c@{\hskip 2pt}c@{\hskip 2pt}c@{\hskip 2pt}c@{\hskip 2pt}c@{\hskip 2pt}c@{\hskip 2pt}c@{\hskip 2pt}c@{\hskip 2pt}c}
\includegraphics[width=0.170\textwidth]{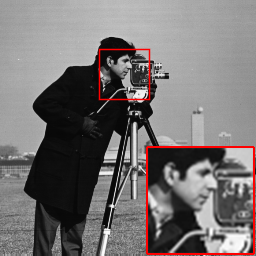}&
\includegraphics[width=0.170\textwidth]{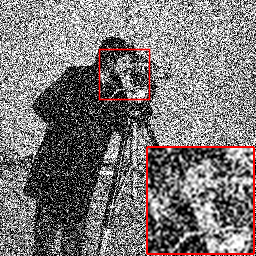}&
\includegraphics[width=0.170\textwidth]{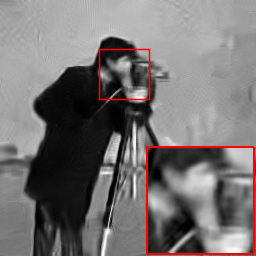}&
\includegraphics[width=0.170\textwidth]{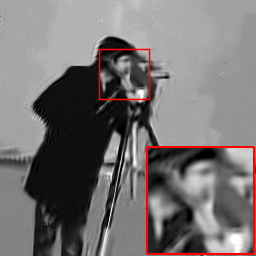}&\vspace{-5pt}\\
\scriptsize (a) Original &
\scriptsize (b) Niosy image &
\scriptsize (c) BM3D\cite{dabov2007image}~\scriptsize{23.07/0.692}&
\scriptsize (d) WNNM\cite{gu2014weighted}~\scriptsize{23.36/0.697}&\\
\includegraphics[width=0.170\textwidth]{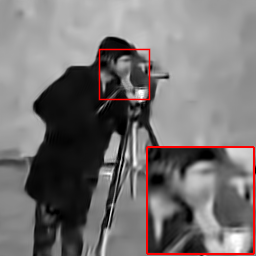}&
\includegraphics[width=0.170\textwidth]{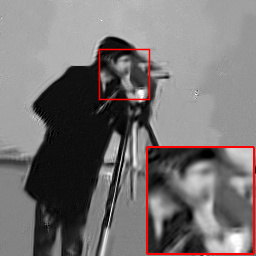}&
\includegraphics[width=0.170\textwidth]{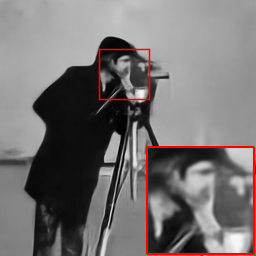}&
\includegraphics[width=0.170\textwidth]{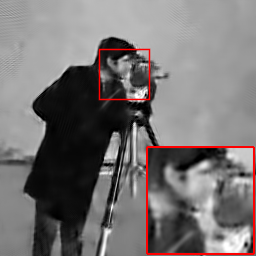}&\vspace{-5pt}\\
\scriptsize (e) PCLR~\cite{chen2015external}~\scriptsize{23.48/0.715}&
\scriptsize (f) OLRA\cite{zhang2018kernel}~\scriptsize{23.32/0.700}&
\scriptsize (g) FFDNet\cite{Zhang2018FFDNet}~\scriptsize{23.96/0.740}&
\scriptsize (h) Ours \scriptsize{23.96/0.721}\\
 \end{tabular}
 \caption{  \small{{Denoising results on image Cameraman by different methods with noise level ~$\sigma  = 100$. There is no restored image of DnCNN in this figure because it is untrained for $\sigma=100$.  } }
}
\label{Fig: man}
\end{figure*}


\section{Experimental results and discussion}
\label{sec 3}

From the 
 BSD dataset~\cite{martin2001database}, 200 training images are uniformly sampled to $2\times10^6$ image patches, where GMM with 250 mixing components and parameter $\bm{\Theta}$ are learned. The related initialization parameters are set to $\alpha=0.10$, $\beta=0.62$,  and the number of iterations $MaxIter=5$.

The test images are selected from the BSD dataset~\cite{martin2001database} (see Figure \ref{Fig testimages}). There are five $256 \times 256$ images and five $512 \times 512$ images. Our algorithm PG-LR is compared with several state-of-the-art algorithms, including BM3D~\cite{dabov2007image}, WNNM~\cite{gu2014weighted}, PCLR~\cite{chen2015external},
OLRA~\cite{zhang2018kernel}, DnCNN~\cite{zhang2017beyond} and FFDNet~\cite{Zhang2018FFDNet}. All source codes are got from the authors.

Quantitatively, we evaluate the quality of image restoration by the peak signal to noise ratio (PSNR) and structural similarity index (SSIM). The SSIM is defined by Wang et al.~\cite{wang2004image} and PSNR is given by
\begin{equation*}
\mathrm{PSNR} =   10\log_{10}\frac{255^2}{\mathrm{MSE}},
\end{equation*}
where
\begin{equation*}
\mathrm{MSE} = \frac{1}{m\times n} \|\mathbf{X}-\widetilde{\mathbf{X}}\|_{F}^{2} ,
\end{equation*}
$m \times n$ is the image size, $\mathbf{X}$ is the original image and $\widetilde{\mathbf{X}}$ is the restored image.

We record the values of PSNR / SSIM of restored image at various Gaussian noise level in Table \ref{tab:table a}. The table confirms that our method achieves the best results in almost all the cases in both PSNR and SSIM values among the traditional denoising algorithms and is also a little better than some deep learning based methods. 
Furthermore, our algorithm wins BM3D~\cite{dabov2007image} more than $ 1 dB$ in average when $\sigma=100$.

For  visual image quality, Figures \ref{Fig: house} and \ref{Fig: man}  show that the proposed method yields the best image visual quality in terms of removing
noise, preserving edges and maintaining image details.
 Taking Figure \ref{Fig: house} for example, the part zoomed in shows that our proposed method obtains pleasurable result at window frame while BM3D blurs the window. The low rank based algorithms (WNNM~\cite{gu2014weighted}, PCLR~\cite{chen2015external} and OLRA~\cite{zhang2018kernel}) introduce unpleasant artifacts, and the deep learning based methods cut the structure at the lower left corner of the window.

When the noise level is extremely high, the structural characteristics of the image are severely damaged and the restoration task becomes much arduous.
In Figure \ref{Fig: man}, the cameraman's ear is apparently restored almost completely by our algorithm, while other methods introduce serious  artifacts in the hair and destroy the ear.
FFDNet~\cite{Zhang2018FFDNet} sharpens the edges but removes some structures partly such as 
the details of the camera.
Though our approach blurs the edges a little bit, it preserves most of the structures.
In summary, our algorithm remains more detail information such as structures and edges than other advanced methods.

In brief, the proposed algorithm has competitive performance of removing noise and remaining details. It outperforms the state-of-the-art methods both quantitatively and in visual quality, including several deep learning methods such as DnCNN~\cite{zhang2017beyond} and FFDNet~\cite{Zhang2018FFDNet} .

\section{Conclusion}
\label{sec 4}

In this paper, we propose a novel method by combining Gaussian patch mixture model and low rank matrix restoration.
We conduct extensive experiments at different Gaussian noise levels, showing that the proposed approach is robust and outperforms the state-of-the-art in image denoising.
The denoising capability of our proposed method largely benefits from the accuracy of similar patches selection, thereby it makes up for the lack of ability for low rank approximation and maximum expectation algorithms to maintain the details and edges.
Future works include:
(1) Global patch matching greatly increases the complexity by spreading over all image 
patches to find their similar ones.
How to reduce the computational complexity and ensure great denoising performance is worth discussing.
(2) The parameters in the simulation experiment are selected according to experience. The slight changes of a parameter will also make difference.
Therefore, how to set an adaptive parameter function is also a point worth studying.
(3) In the process of experiments, we find that when performing the classification operation, some classes have only a few similar patches while some others get so many ones.
It is rational to set appropriate classes to balance the number of similar patches in each class, thereby enhancing the robustness of the algorithm.
(4) The future work also includes extensions to other types of noise such as Poisson noise, mixed noise and unknown type of noise.

\section*{Acknowledgments}
{
Jin has been supported by the National Natural Science Foundation of China (Grant No. 12061052), Natural Science Fund of Inner Mongolia Autonomous Region (Grant No. 2020MS01002), China Scholarship Council for a one year visiting at Ecole normale sup\'{e}rieure Paris-Saclay (No. 201806810001).
}

{\small
\bibliographystyle{ieee_fullname}
\bibliography{01egbib}
}

\end{document}